\documentclass[10pt,twocolumn,letterpaper]{article}

\usepackage[pagenumbers]{wacv} %

\usepackage{graphicx}
\usepackage{amsmath}
\usepackage{amssymb}
\usepackage{booktabs}
\usepackage{etoolbox}
\usepackage{enumitem}
\usepackage{xcolor}
\usepackage[accsupp]{axessibility}

\usepackage[pagebackref,breaklinks,colorlinks]{hyperref}

\usepackage[capitalize]{cleveref}
\crefname{section}{Sec.}{Secs.}
\Crefname{section}{Section}{Sections}
\Crefname{table}{Table}{Tables}
\crefname{table}{Tab.}{Tabs.}

\def\wacvPaperID{2448} %
\def\confName{WACV}
\def\confYear{2024}

\providetoggle{showcomments}
\settoggle{showcomments}{false}

\iftoggle{showcomments}{%
    \newcommand{\todo}[1]{\textcolor{blue}{\textbf{TODO:} #1}}
    
    \newcommand{\new}[1]{\textcolor{purple}{\textbf{NEW:} #1}}
    \newcommand{\michael}[1]{\textcolor{orange}{\textbf{Michael:} #1}}
    \newcommand{\matteo}[1]{\textcolor{red}{\textbf{Matteo:} #1}}
}{%
    \newcommand{\todo}[1]{}
    
    \newcommand{\new}[1]{#1}
    \newcommand{\michael}[1]{}
    \newcommand{\matteo}[1]{}
}

\newcommand{\para}[1]{\par\vspace{0pt}\noindent\textbf{#1}}

\newcommand{\uc}{\expandafter\MakeUppercase}

\begin{document}

\title{CycleCL: Self-supervised Learning for Periodic Videos}
\author{Matteo Destro\\
Cerrion AG\\
{\tt\small matteo@cerrion.com}
\and
Michael Gygli\\
Cerrion AG\\
{\tt\small michael@cerrion.com}
}
\maketitle

\begin{abstract}
Analyzing periodic video sequences is a key topic in applications such as automatic production systems, remote sensing, medical applications, or physical training. An example is counting repetitions of a physical exercise.
Due to the distinct characteristics of periodic data, self-supervised methods designed for standard image datasets do not capture changes relevant to the progression of the cycle and fail to ignore unrelated noise. They thus do not work well on periodic data.

In this paper, we propose CycleCL, a self-supervised learning method specifically designed to work with periodic data.
We start from the insight that a good visual representation for periodic data should be sensitive to the phase of a cycle, but be invariant to the exact repetition,
\ie it should generate identical representations for a specific phase throughout all repetitions.
We exploit the repetitions in videos to design a novel contrastive learning method based on a triplet loss that optimizes for these desired properties. Our method uses pre-trained features to sample pairs of frames from approximately the same phase and negative pairs of frames from different phases. Then, we iterate between optimizing a feature encoder and re-sampling triplets, until convergence.

By optimizing a model this way, we are able to learn features that have the mentioned desired properties. We evaluate CycleCL on an industrial and multiple human actions datasets, where it significantly outperforms previous video-based self-supervised learning methods on all tasks.

\end{abstract}

\section{Introduction}
\label{sec:intro}
\begin{figure*}[t]
  \centering
  \includegraphics[width=0.85\linewidth]{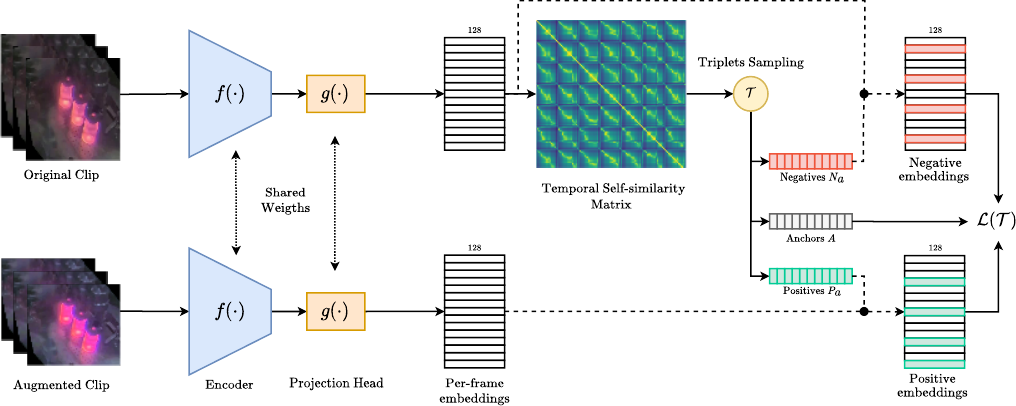}
  \caption{\textbf{Overview of our SSL method for Periodic Videos:} frames are encoded with a CNN and used to build a Temporal Self-Similarity Matrix (TSM) between the frames in a clip. From this matrix, we sample positive and negative frames for each anchor by selecting the most similar or dissimilar frames to build triplets. We then minimize a triplet loss, thereby pulling positives closer to the anchor and pushing negatives away. We iterate this process until convergence. To make the model more robust, we augment the positives with a set of random transformations and use them as inputs when computing and minimizing the loss.
  }
  \label{fig:architecture_cyclecl}
\end{figure*}
When analyzing temporal data, repetitions are omnipresent in most types of applications:
repetitions of physical exercises~\cite{rep_dataset_countix};
discrete sets of repeating steps in automatic production processes~\cite{dl_for_manufacturing3};
seasonal changes in remote sensing ~\cite{verbesselt2010detecting};
periodic patterns in medical applications, where they play a key role when analyzing vital signs such as heart rate or respiration ~\cite{kossack2019local,bae2022prospective}.

Thus, feature representations that capture the main aspects of periodic data are of key interest.
In most cases, such data is very different from what can be found in standard datasets like ImageNet~\cite{dataset_imagenet} or the Kinetics dataset~\cite{carreira2017quo}, which are visually and semantically extremely diverse. Instead, in periodic data diversity is limited, and small differences are often fundamental. For instance, satellite images might always capture the same region with the goal of finding when seasonal patterns deviate from normality, \eg during a drought. For models to perform well on such data, they need to be able to detect subtle changes. This limits the success of the de-facto standard of employing transfer learning from existing datasets~\cite{transfer_learning_imagenet,transfer_learning_decaf,transfer_learning_visual_adaptation_benchmark, transfer_learning_michaels}.
At the same time, labeled data for these applications is typically scarce and hard to acquire. Thus, in this paper we propose to employ self-supervision to learn strong representations for periodic videos. These features can then be used for a set of downstream tasks such as repetition counting~\cite{rep_dataset_countix} or anomaly detection. Focusing on feature learning is motivated by works such as~\cite{panda}, which showed that with sufficiently general representations, a simple model like k-nearest neighbors achieves strong performance on tasks such as anomaly detection.

A plethora of self-supervised methods have been proposed to tackle general feature learning, \eg~\cite{gidaris2018unsupervised, info_nce_loss, simclr, byol, mocov2, tcc, lav}. These methods, however, are typically designed for datasets such as ImageNet and struggle with periodic data. The prevalent contrastive methods~\cite{info_nce_loss}, such as SimCLR~\cite{simclr}, use random augmentations of the anchor as positive samples, and randomly sampled images as negative samples.
Therefore, the model is trained to be invariant to pre-defined transformations of the \emph{same} instance. 
This approach works well on highly diverse datasets, but becomes less effective on periodic data,
as evidenced in Sec.~\ref{sec:feature_quality}, where a nearly identical image may be repeated once every cycle.
In such cases, randomly selecting negative samples and forcing the representation to be dissimilar impedes the learning process,
making a smart selection mechanism necessary.

\begin{figure*}[t]
    \centering
    \includegraphics[width=0.95\linewidth]{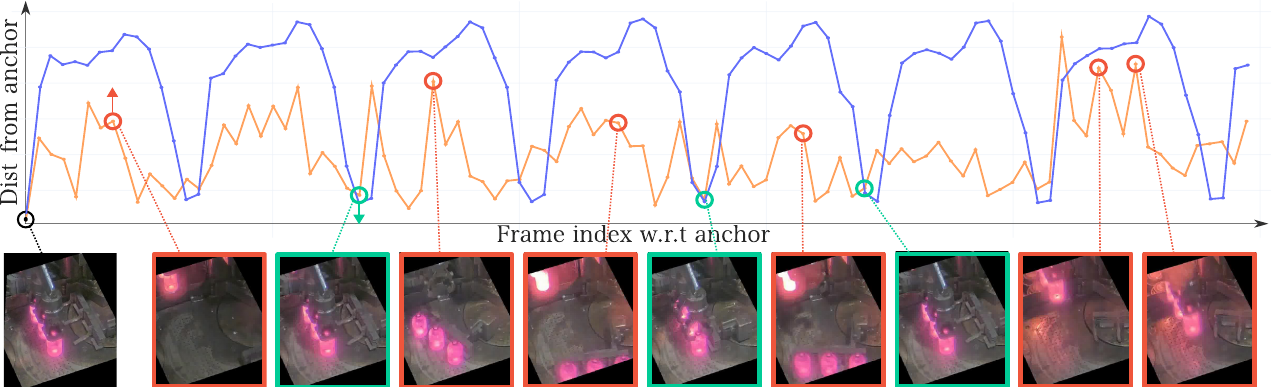}
    \caption{%
    \definecolor{green}{HTML}{00cc96}%
    \definecolor{red}{HTML}{ef553b}%
    \definecolor{orange}{HTML}{ffa15d}%
    \definecolor{blue}{HTML}{636efa}%
    \textbf{Our triplet selection mechanism:} the feature similarity between consecutive frames is exploited to estimate a set of triplets such that the positives ({\color{green}green}) belong to the same phase as the anchor ({\color{black}black}) but in successive repetitions, while the negatives ({\color{red}red}) belong to different phases. At the beginning of the training process the similarity ({\color{orange}orange line}) is noisy and the selection is inaccurate, but after just five epochs ({\color{blue}blue line}) the model learns representations that are sensitive to the phase but invariant to the specific repetition.
    }
    \label{fig:cyclecl_selection_full}
\end{figure*}

We start from the following insight: a good visual representation for periodic data should be sensitive to the progression/phase in the cycle. At the same time, it should be invariant to the exact repetition,~\ie produce the same representation for the same phase across cycles.
We exploit this property to design our novel self-supervised learning method with smart triplet sampling, which is trained via a triplet loss~\cite{triplet_loss}.
Specifically, we hypothesize that even with a suboptimal feature representation it is possible to find frames that are clearly showing a different phase of the cycle. At the same time, it should be possible, with better than random accuracy, to find pairs of frames showing roughly the same phase. We use this idea to compute similarities between frames and, from these, sample initial positive and negative pairs, and start the optimization process. Our method iterates between computing new triplets and optimizing the encoder, until convergence. This is similar to deep 
clustering~\cite{caron2018deep}, which alternates between assigning images to clusters and optimizing the feature representation by predicting the cluster index of an image.
In our experiments, we show that by optimizing the feature representation this way, we are able to learn frame embeddings that have the desired properties described above (see Fig.~\ref{fig:cyclecl_selection_full}).

Concretely, we make the following contributions:
(i) A novel self-supervised learning method designed for periodic video data and applicable over a wide range of domains.
(ii) An augmentation strategy adapted to periodic data, further boosting the performance of our method.
(iii) Evaluation on an industrial and on three human actions dataset over two tasks.
Our approach outperforms previous self-supervised learning methods %
such as SimCLR~\cite{simclr}, RepNet~\cite{rep_dataset_countix}, Temporal Cycle Consistency~\cite{tcc} and DINO~\cite{dino}
on nearest neighbor classification and unsupervised anomaly detection.
On the indutrial dataset, for example, our method reduces the gap to a fully supervised method to 8\%, from 18\% for the second best method (TCC).
The remainder of the paper is structured as follows:
in Sec.~\ref{sec:related_work} we discuss related work. Sec.~\ref{sec:method} introduces our method. Sec.~\ref{sec:experiments} experimentally validates the benefits of our approach. Finally, Sec.~\ref{sec:conclusions} concludes our paper.

\section{Related Work}
\label{sec:related_work}
Our method relates to existing works for self-supervised representation learning, in particular to methods that learn representations from videos via temporal signals.

\para{Self-supervised learning (SSL).}
This is an effective way for pre-training deep models without the need for human supervision. SSL is based on the idea that a model can learn general visual representations by solving a \emph{pretext task}. Such a task uses pseudo-labels extracted programmatically from the data itself, rather than human-provided labels. Through using a representation learned this way, little to no supervision is necessary to solve a \emph{downstream task}, such as anomaly detection or image classification. 
An early canonical example of SSL is based on rotating images and predicting their rotation angle~\cite{gidaris2018unsupervised}. Recent methods use contrastive learning~\cite{simclr, byol, he2020momentum, mocov2}, a form of metric learning, where positive pairs are created based on augmentations of an image itself.
\cite{tian2021divide} proposes an extension to contrastive methods, tailored to diverse and heavy-tailed image datasets. Via clustering the dataset into semantically-similar subsets, they create relevant hard negatives by sampling within these clusters. When learning on repetitive videos we encounter the opposite problem, making this method unsuitable for periodic data (see Sec.~\ref{sec:intro})

An alternative to contrastive approaches is DINO~\cite{dino}, a self-supervised method for vision transformers~\cite{dosovitskiy2020image}, and its extensions~\cite{oquab2023dinov2,ranasinghe2022self}. DINO is a form of knowledge distillation with no labels, where the student is trained to predict the same features as the teacher, whose weights are the moving average of the student's weights, rather than directly optimized weights.
Recently, masked autoencoders (MAE)~\cite{mae} have been proposed for feature learning. They also rely on vision transformers~\cite{dosovitskiy2020image}, but instead aim to reconstruct a masked input. While generally performing well, MAE models are worse in anomaly detection tasks than previous self-supervised methods, as shown by Reiss~\etal \cite{ssl_for_ad}.

Apart from these most relevant works, a plethora of other SSL methods exist. For a more in-depth review of self-supervised learning methods, please refer to~\cite{ssl_survey_1, ssl_survey_2, Balestriero2023}.

\para{Learning representations from videos.}
Several methods rely on the temporal dimension of videos to design self-supervised objectives~\cite{jenni2020video,tcc,lav,rep_count_align_sampling, rep_dataset_countix,qian2021spatiotemporal}. Temporal Cycle Consistency (TCC) \cite{tcc} proposes an objective over pairs of videos. It optimizes the representation such that if a frame \emph{b} in one video is the nearest neighbor for frame \emph{a} in another video, the inverse holds as well. \cite{kong2020cycle} extends \cite{tcc} by additionally maximizing the similarity between frame and video representations (which are average pooled frame representations).
Instead of using separate local and global objectives,
\cite{lav} learns a representation by globally aligning video pairs using a differentiable version of Dynamic Time Warping~\cite{dtw_ts}. Both methods however require class labels for each video to build pairs and struggle with periodic data, since the alignment becomes ambiguous.

More related,~\cite{rep_count_align_sampling, rep_count_simper, rep_dataset_countix} also learn from periodic videos, specifically to perform repetition counting. SimPer~\cite{rep_count_simper} is based on contrastive prediction similar to SimCLR~\cite{simclr}. Instead, RepNet~\cite{rep_dataset_countix} is trained through creating synthetic repetitions and using the period length as a supervision signal.
Jacquelin~\etal~\cite{rep_count_align_sampling} sample triplets from periodic videos to optimize a metric learning objective, as in our work. However, they use a fixed sampling scheme relying on adjacent frames. Our method instead dynamically samples frames across different repetitions within the same video, thus producing more general features.
This leads to more powerful representations, as they are invariant to the exact repetition of a periodic input, while increasing the sensitivity to the progression in a period.

In our experiments (Sec.~\ref{sec:experiments}) we compare our approach to~\cite{simclr,dino,tcc,rep_count_align_sampling,rep_dataset_countix}, \new{covering multiple canonical categories of methods designed to address this problem. Our findings show that CycleCL outperforms these alternatives on periodic video datasets.}

\section{Our method: CycleCL}
\label{sec:method}
We introduce CycleCL, a self-supervised learning method for learning feature representations on periodic data (Fig.~\ref{fig:architecture_cyclecl}).
Periodic video or image sequences have a set of key characteristics:
(i) They exhibit a repetitive signal.
(ii) Variations during a repetition might be visually subtle, ~\eg when alternatingly lifting a foot during a planking exercise.
(iii) Variations irrelevant to the repetition might be prominent,~\eg when the camera is not static, there are camera flashes or partial occlusions of the repeating pattern, like in Fig.~\ref{fig:countix_examples}.
Therefore, an effective visual representation for periodic data should accurately capture the phase of a cycle while remaining unaffected between repetitions. In other words, it should yield the same representation for a given phase across all repetitions. 
Our novel self-supervised learning method exploits the periodicity in videos to directly optimize a feature encoder to have these properties. 
Our method iterates between:
\begin{enumerate}[label={(\roman*)},leftmargin=18pt,itemsep=2pt]
\item Sampling triplets: for each frame, finding other frames that represent the same phase but in a different repetition (positives), and frames that depict a different~phase~(negatives).
\item Optimizing the encoder \( f(\cdot) \),
such that the distance from the anchor to the positives becomes smaller compared to the distance from the anchor to the negatives.
\end{enumerate}
By iterating this process, our method learns features that capture the variations that are relevant to the periodic signal and ignore other irrelevant changes.

To optimize the encoder we specifically minimize a \emph{triplet loss} \cite{triplet_loss} during training:
\begin{multline}
    \mathcal{L}(\mathcal{T}) = \sum_{(v_a, v_p, v_n) \in \mathcal{T}} \bigg[ \Vert f(v_a) - f(v_p) \Vert^2_2 \\ - \Vert f(v_a) - f(v_n) \Vert^2_2 + \alpha \bigg]_+
\label{eq:loss}
\end{multline}
where \( \mathcal{T} \) is the set of valid triplets \( (v_a, v_p, v_n) \). The margin parameter \( \alpha\) guarantees that only (semi-)hard samples play a role in the training process. Triplets where the negative sample is farther from the anchor than the positive sample by more than \( \alpha\) do not contribute to the loss.
\new{We select triplet loss as it operates on a specific set of positive and negative examples. In contrast, more recent loss functions, such as \emph{InfoNCE} \cite{info_nce_loss}, use a single positive and treat all other samples in a batch as negatives, which leads to an incorrect labeling for periodic data.}

Next, we discuss our triplet sampling strategy and network architecture in detail.

\begin{figure}[t]
    \centering
    \begin{subfigure}[t]{1\linewidth}
        \centering
        \includegraphics[width=0.95\linewidth]{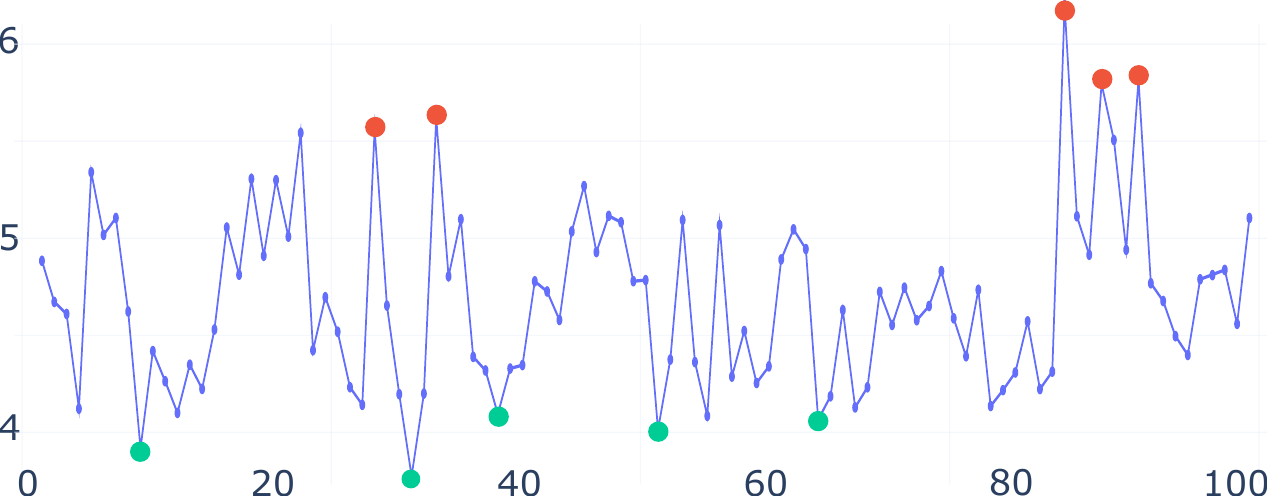}
        \caption{Min/Max Top-k}
        \label{fig:cyclecl_selection_topk}
    \end{subfigure}%
    \hfill
    \begin{subfigure}[b]{1\linewidth}
        \centering
        \includegraphics[width=0.95\linewidth]{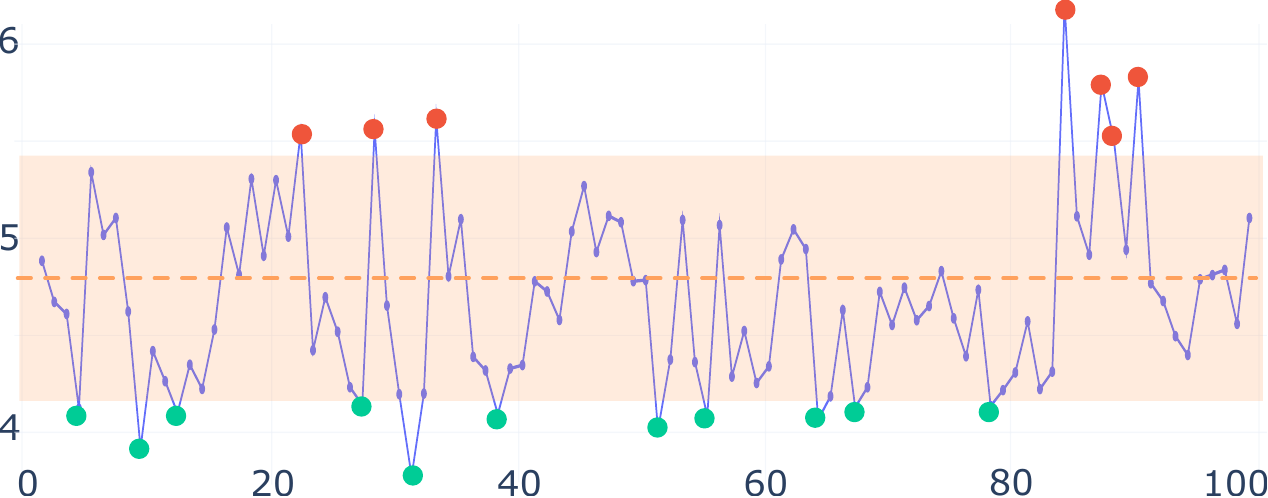}
        \caption{Mean Thresholding}
        \label{fig:cyclecl_selection_mean}
    \end{subfigure}
    \caption{Visualization of the proposed triplets sampling methods. The green and red dots represent, respectively, the selected positive and negative samples for the current frame at index 0.}
    \label{fig:cyclecl_selection_methods}
\end{figure}
\para{Triplet sampling.}
\label{subsubsec:sampling_methods}
When training with the loss in Eq.~\eqref{eq:loss}, the key is sampling adequate triplets.
The goal is to select frames as positives if they depict the same phase as the anchor and as negatives otherwise.
Devoid of ground truth, this can not initially be done perfectly. But even with an unoptimized feature encoder, it is possible to find such samples with better than random accuracy~\cite{caron2018deep}.
We do this by sampling positive and negative pairs based on the similarity of their feature representations.
Initially, these features are produced by a model trained for ImageNet classification and a randomly initialized head, but as the encoder \( f(\cdot) \) gets updated in consecutive steps, the feature representations evolve and allow more accurate sampling.

Fig.~\ref{fig:cyclecl_selection_full} shows how our sampling strategy provides a relevant learning signal.

\begin{figure}[t]
    \centering
    \hfill
    \begin{subfigure}[b]{.46\linewidth}
        \centering
        \includegraphics[width=1\linewidth]{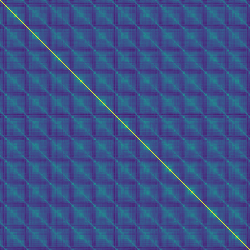}
        \includegraphics[width=1\linewidth]{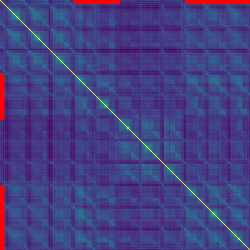}
        \caption{ImageNet pre-trained}
    \end{subfigure}
    \hfill
    \begin{subfigure}[b]{.46\linewidth}
        \centering
        \includegraphics[width=1\linewidth]{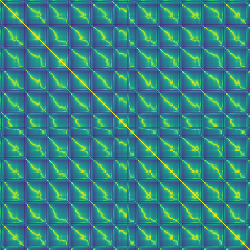}
        \includegraphics[width=1\linewidth]{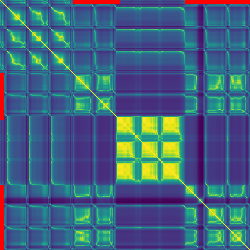}
        \caption{CycleCL (Ours)}
    \end{subfigure}
    \hfill
    \caption{Temporal Self-similarity Matrices (TSM) produced with a model trained on ImageNet and with CycleCL.
    A lighter color represents a higher similarity.
    In the video depicting a \emph{periodic} process (top), the embeddings from the ImageNet baseline do not show strong similarity patterns at regular intervals. The model is not invariant to insignificant changes in the input.
    Our CycleCL model instead captures the periodicity accurately.
    For the \emph{non-periodic} video (bottom), the TSM from CycleCL clearly shows a difference in similarity when an anomaly occurs (marked by the red lines), while the ImageNet model is not able to capture this change as clearly.
    }
    \label{fig:tsm_examples}
\end{figure}

The prerequisite to sampling triplets is a temporal self-similarity matrix (TSM). Given a representation \( f(v_i) \) for each frame \( v_i \) of the input video, we compute the self-similarity matrix \(\mathcal{S} \) using the \emph{negative squared Euclidean distance}.
Since the input videos can be very long, and we are interested in modeling short-term cycles, \( S \) is computed only on a \emph{chunk} of the video, i.e. a window of \( C \) subsequent frames. The result is a matrix with shape \( |C| \times |C| \), where \( \mathcal{S}_{ij} \) is the similarity between the embeddings of frames \( v_i \) and \( v_j \), computed as:
\begin{equation}
    \label{eq:tsm}
    \mathcal{S}_{ij} = -\mathcal{D}_{ij} = -\Vert f(v_i) - f(v_j) \Vert^2_2
\end{equation}
An example of self-similarity matrix is shown in Fig.~\ref{fig:tsm_examples}.
We propose two different sampling strategies based on the similarity matrix \( \mathcal{S} \): 

\textit{(i) Min/Max Top-k}: given an anchor frame \( v_a \) from a video chunk \( C \), the set of positive frames \(P_a\) is sampled using the \( k \) \emph{most similar} frames, \ie \(P_a = \operatorname*{\emph{k}-argmax}_{v_p \in C} \mathcal{S}_{ap}\), while the negatives \(N_a\) are taken from the \( k \) \emph{most dissimilar} ones.

\textit{(ii) Mean Thresholding}: the mean similarity between the anchor \( v_a \) and all the frames in \( C \) is used as threshold for the selection, computed as: \( \mu_a = \frac{1}{|C|} \sum_{v \in C} \mathcal{S}_{av}  \),
where \( \mathcal{S}_{av} \) is the feature similarity between the anchor and video \( v \). Frames with similarity below the threshold are taken as positives and above as negatives. An additional hyperparameter \( \beta \) controls the selection margin between the positives and the negatives, to ensure that the selected frames only have a strong similarity or dissimilarity. The final sets of positives and negatives are defined as:
\begin{align}
    P_a &= \{ v_p \in C \mid \mathcal{S}_{ap} > \mu_a (1 + \beta) \} \\
    N_a &= \{ v_n \in C \mid \mathcal{S}_{an} < \mu_a (1 - \beta) \}
\end{align}

Fig.~\ref{fig:cyclecl_selection_methods} presents a visual comparison of the selection methods.
While the number of positives and negatives is a fixed hyperparameter for \emph{Min/Max Top-k}, their number is automatically set in  \emph{Mean thresholding} based on relative similarity, allowing it to select more accurate triplets.

\para{Frame Encoder.}
The frame encoder \( f(\cdot) \) is composed of two main components:

\textit{(i) Feature Extractor}: each frame \(v_i\) is fed separately to a convolutional encoder to extract intermediate embeddings. The spatial dimensionality is kept. We used a ResNet model in our experiments, but this component is architecture-agnostic.

\textit{(ii) Projection Head}: The convolutional features undergo dimension reduction through a 3D convolutional layer, global max pooling, and a linear layer. The 3D convolution helps the network to model short-term temporal patterns to better distinguish similar-looking frames but with different motion contexts.

\para{Augmentations.}
We propose to augment the positive frames with random image transformations when optimizing the model.
As the triplet loss (Eq.~\eqref{eq:loss}) forces the positives to be closer to the anchor than the negatives by some margin, adding augmentations on the positives, but not the negatives, makes the task harder by forcing the network to become invariant to these transformations and thus makes it more robust~\cite{shorten2019survey}. The augmentations are used only when computing the loss. The triplets selection is performed on the unaltered frames to avoid changing the TSM and thus reducing the quality of the sampled triplets.

\section{Experiments}
\label{sec:experiments}
\begin{table*}[t]
    \centering
    \begin{tabular}{l|c|c|c|c|c|c|c|c}
    \hline
     & \multicolumn{2}{c|}{\textbf{Industrial}} & \multicolumn{2}{|c}{\textbf{Countix}} & \multicolumn{2}{|c}{\textbf{QUVA}} & \multicolumn{2}{|c}{\textbf{PERTUBE}} \\
    \textbf{Method} & \textbf{AP} & \textbf{\( F_1 \)} & \textbf{AP} & \textbf{\( F_1 \)} & \textbf{AP} & \textbf{\( F_1 \)} & \textbf{AP} & \textbf{\( F_1 \)}\\
    \hline
    Random & 04.05 & 03.95 & 36.62 & 30.67 & 06.07 & 02.06 & 15.95 & 11.42 \\
    ImageNet pre-trained & 46.25 & 46.31 & 40.23 & 37.18 & 10.30 & 10.12 & 27.89  & 27.05 \\
    \hline    
    SimCLR~\cite{simclr} & 27.45 & 29.62 & 39.12 & 38.84 & 12.38 & 09.24 & 30.45 & 29.51 \\
    RepNet~\cite{rep_dataset_countix} & - & - & 41.24 & \textbf{43.22} & 12.88 & 12.85 & 35.39 & 32.22 \\    
    DINO~\cite{dino} & 50.65 & 50.92 & 38.24 & 37.40 & 12.47 & 12.03 & 33.45 & 32.59 \\
    TCC~\cite{tcc} & 53.88 & 54.14 & 39.76 & 39.41 & 11.42 & 11.61 & 31.29 & 30.61 \\
    \hline
    \hline
    CycleCL (Ours) & \textbf{61.35} & \textbf{63.86} & \textbf{45.27} & 43.16 & \textbf{13.79} & \textbf{13.65} & \textbf{36.98} & \textbf{37.47} \\
    \hline
    \end{tabular}        
    \caption{Comparison against baselines and current state-of-the-art approaches. Our CycleCL method significantly outperforms the ImageNet baseline and previous state-of-the-art methods. Notably, it significantly outperforms the canonical SimCLR~\cite{simclr} which struggles with periodicity, where nearly identical images are repeated once every cycle. In such cases, randomly selecting negative samples, as done in this method, is suboptimal.
It also compares favorably with RepNet \cite{rep_dataset_countix}, which is specifically designed for periodic data.
}    
\label{tab:sota_methods_comparison}
\end{table*}
In this section, we compare our approach with previous state-of-the-art methods. We arrange the evaluation in two steps, namely \emph{feature learning} and \emph{anomaly detection}, and focus on their results separately to get a better overview of the single contributions.

\para{Datasets.}
We evaluate our method on four diverse datasets. The first one, which we refer to as \emph{Industrial} dataset, consists of videos from production lines in the glass packaging industry, which we collected. This dataset consists of 575 unlabeled videos for training and 707 labeled videos for testing, with an average duration of 2 min. We collected these samples from cameras installed in production lines, and annotated anomalous events leading to loss of products.

\begin{figure}[t]
    \centering
    \includegraphics[width=0.26\linewidth]{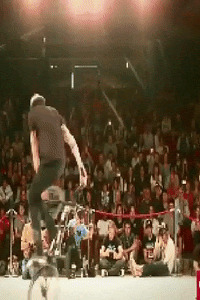}
    \includegraphics[width=0.26\linewidth]{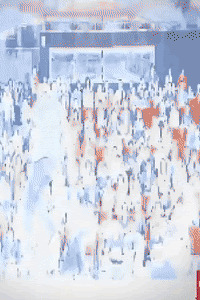}
    \includegraphics[width=0.26\linewidth]{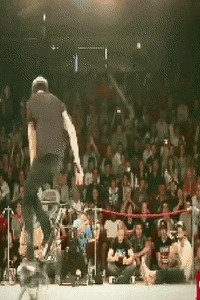}
    \caption{Example from ~Countix~\cite{rep_dataset_countix}.~Even for frames showing the same phase of a bike spin, there is large variation due to position differences, camera zoom, illumination changes and a camera flash, highlighting the difficulty of learning relevant features.}
    \label{fig:countix_examples}
\end{figure}

The second is \emph{Countix}~\cite{rep_dataset_countix}, a subset of \emph{Kinetics}~\cite{carreira2017quo} containing repetitive actions usually used as a benchmark for repetition counting tasks (examples in Fig.~\ref{fig:countix_examples}).

To ensure a comprehensive evaluation, we also include the \emph{QUVA}~\cite{rep_dataset_quva} and \emph{PERTUBE}~\cite{rep_dataset_pertube} datasets in our evaluation. These datasets are closely related to Countix in terms of domain but are significantly smaller in size. QUVA is composed of 100 videos, while PERTUBE consists of only 50 videos. Given that, we follow \cite{rep_dataset_countix} and use them as test datasets on which we evaluate the models trained on Countix.

All the datasets have labeled intervals from two classes: \emph{periodic} and \emph{non-periodic}. For the Industrial dataset, the samples were manually annotated by an expert. For Countix, QUVA and PERTUBE the labels were provided by the authors.

\para{Implementation details.}
We employ a ResNet-18 \cite{resnet} model as encoder, pre-trained on ImageNet. The outputs of the last convolutional layer are stacked along the temporal dimension in chunks of 64 frames and forwarded to the projection head, followed by \( L_2 \) normalization. During training, we train to convergence,~\ie we stop the training process when the loss drops to zero for multiple consecutive iterations, \new{which typically happens after around 40 epochs.}
We use the Adam optimizer, with a learning rate of \( 10^{-4} \). We sample long video clips of 100 frames, to maximize the number of valid triplets at each iteration. More details in the supp. material.

\subsection{Feature Quality}
\label{sec:feature_quality}
We first evaluate the robustness of the learned representations in terms of separation between \emph{non-periodic} and \emph{periodic} samples.

\para{Evaluation protocol.}
We evaluate our method on the four datasets and compare it against previous approaches discussed in Sec.~\ref{sec:related_work}.
In line with other recent works~\cite{dino, knn_eval}, we use a weighted k-NN classifier on top of the features \( f(x_i) \) to determine if a frame is taken from an interval depicting a periodic process or not. 
Since video data has a strong temporal redundancy between neighboring frames, we do leave-one-out evaluation,~\ie we exclude frames belonging to the same video of \( x_i \) when computing the nearest neighbors.
The set \( N_k \) of top \( k \) nearest neighbors is determined using the Euclidean distance. The final classification is computed via weighted voting. Following the evaluation setting of Wu et al. \cite{knn_eval}, a class \( c \) gets a total weight \( 
    w_c = \sum_{i \in N_k} \alpha_i \mathbf{1}_{c_i = c} \), where \( \alpha_i = \frac{1}{d_{i}^2 + \epsilon}\) and
\( d_{xi} \) is the distance between the target sample \( x \) and sample \( i \). We report \emph{\( F_1 \)-score}, \ie the harmonic mean between recall and precision, as well as \emph{Average Precision} (AP).

\para{Compared methods.}
We compare against SimCLR \cite{simclr}, DINO \cite{dino}, Temporal Cycle Consistency (TCC) \cite{tcc}, and RepNet \cite{rep_dataset_countix}, all state-of-the-art approaches for self-supervised learning. We further include a \emph{random} baseline and a baseline with features of a ResNet-18 model trained on \emph{ImageNet}.

For a fair comparison, all methods use the same
3D backbone as ours and were trained on the same data. 
The only exception is RepNet, where we use the original model trained on Countix, as released by the authors, since the training code was not made available. 
Moreover, since the Industrial dataset significantly differs from Countix, RepNet results for it are not included.

\para{Results (Tab.~\ref{tab:sota_methods_comparison}).} Our method outperforms the current state of the art on all four datasets on both metrics, except RepNet on Countix, where CycleCL outperforms RepNet in AP and is on par in terms of \( F_1 \).
On the industrial dataset, the \( F_1 \) score is improved by 17\% compared to the ImageNet pre-trained baseline, and by 13\% compared to the DINO model.

In terms of previous methods, SimCLR does not perform well on Countix and on the Industrial dataset as it struggles with periodic data, as discussed in Sec.~\ref{sec:intro}.
Optimizing for \emph{Temporal Cycle Consistency}~\cite{tcc} performs well on the Industrial dataset, but worse than SimCLR\cite{simclr} and RepNet\cite{rep_dataset_countix} on the other datasets, especially so on Countix. Countix depicts more diverse actions with larger visual variations, and TCC struggles with repetitions as aligning across videos becomes ambiguous in this case.
RepNet's performance is only slightly worse than our method. But it also requires a more complex augmentation pipeline for generating synthetic videos compared to ours.
To conclude, these strong results show that CycleCL is able to produce robust representations.

Apart from a comparison against other SSL methods, we also evaluated a \emph{supervised} baseline on the industrial dataset. There, the features are extracted from a ResNet-18 model, trained with a supervised signal on the available annotations of the training split. This model achieved 70.81 of Average Precision and 71.91 of 
\(F_1\)-score.
While the performance gap \wrt{} a supervised model persists, our method reduces it to 8\%, compared to more than 18\% for all the other methods.

\begin{figure}[t]
    \centering
    \includegraphics[trim={0 0 6cm 0},clip,width=1\linewidth]{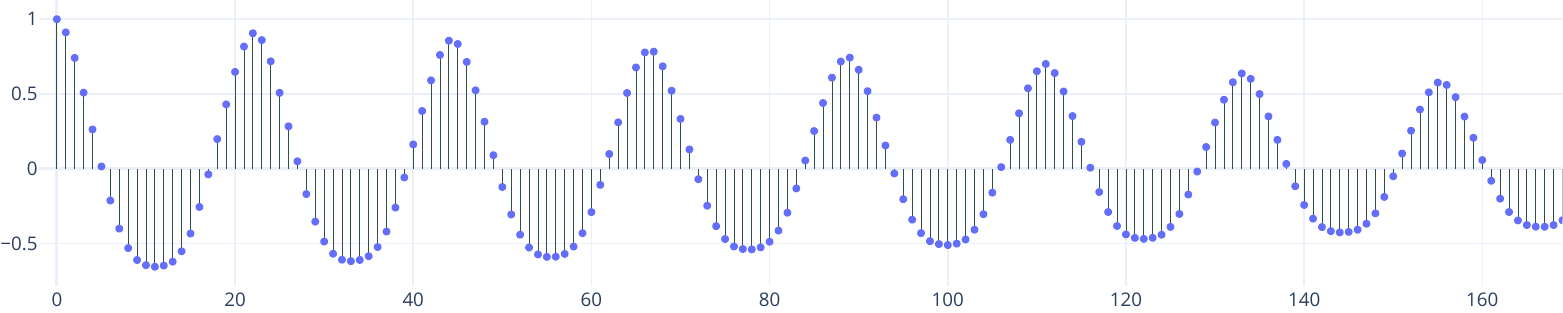}
    \includegraphics[trim={0 0 6cm 0},clip,width=1\linewidth]{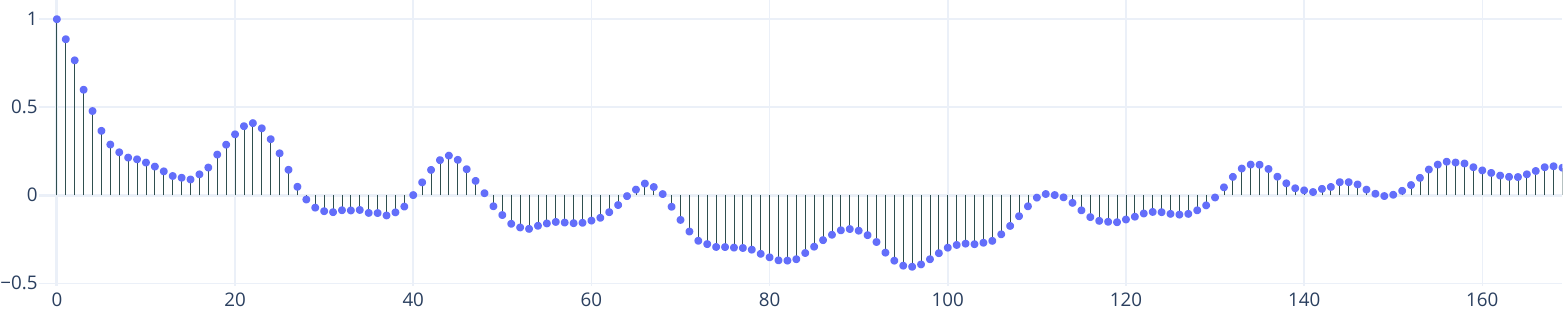}
    \caption{Autocorrelation plot of a periodic (top) and non-periodic (bottom) video. The autocorrelation of two periodic signals measures their correlation as they shift in time, revealing any repeating patterns between them. The cyclic nature of the input is clearly visible during a normal process but disappears during an anomaly.}
    \label{fig:autocorr_examples}
\end{figure}

\para{Visualizations.} Fig.~\ref{fig:tsm_examples} shows an example of TSMs extracted from a periodic and non-periodic video. Our CycleCL model encodes the progression of the cycles better than the ImageNet baseline. Fig.~\ref{fig:autocorr_examples} depicts two autocorrelation plots from the same videos. This visualization shows how good our features are: the period of the cycles can be easily determined by using simple techniques like Fourier transforms.

\subsection{Ablation study}
We perform a set of ablation studies on the Industrial dataset to justify some of the decisions we made when designing CycleCL and show the robustness of our method \wrt{} specific parameter values. Here, we focus on the key design choices (more ablation experiments are provided in the supp. material).

\para{Triplets sampling strategy.} We compare our sampling against other strategies found in the literature (Tab.~\ref{tab:cyclecl_ablation_sampling}). Our \emph{Mean thresholding} approach with \( \beta = 0.3 \) outperforms all the other methods.
Compared to the best-competing strategies proposed in~\cite{tcc, rep_count_align_sampling}, its  \textbf{\( F_1 \)} score is 9\% higher. 

Building triplets by using the nearest frame as positive and the second nearest as negative, as done in the \emph{Adjacent}~\cite{rep_count_align_sampling} method, is not effective on these datasets. Depending on the framerate, the next and the second next frames could have high similarity, and learning to separate them leads to the model focusing on less relevant local changes. 

\new{
\emph{Min/Max Top-k} sampling consistently chooses an equal amount \( k \) of positive and negative samples from each video clip, irrespective of how similar they are to the anchor. Therefore, it does not account for potential differences in the distribution of valid positives and negatives across various clips.
In contrast, \emph{Mean thresholding} does not rely on a fixed \( k \) value, but instead takes into account the similarities to determine the number of triplets to sample. This dynamic approach allows for adjustments based on the varying distributions and the relative balance of positive and negative samples across clips (e.g. a video tends to have more negative pairs than positives). As a result, it leads to a more effective and less noisy selection process.
}

\begin{table}[t]
    \begin{center}
    \begin{tabular}{l|c|c}
    \hline
    \textbf{Method} & \textbf{AP} & \textbf{\( F_1 \)} \\
    \hline
      {Random features} & { 04.05} & {03.95} \\
      {ImageNet features} & {46.25} & {46.31} \\
      Adjacent \cite{rep_count_align_sampling} & 41.32 & 41.65 \\
      TCC~\cite{tcc} & 53.88 & 54.14 \\
      \hline
      Min/Max Top-k (Ours) & 53.76 & 57.54 \\
      Mean Thresholding (Ours) & \textbf{61.35} & \textbf{63.86} \\
      \hline
    \end{tabular}
    \end{center}
    \caption{Ablation study on the triplets sampling strategy for the Industrial dataset. Our best sampling strategy (Mean Thresholding) outperforms the other strategies and baselines.
    }
    \label{tab:cyclecl_ablation_sampling}
\end{table}

\para{Data augmentations.}
When minimizing the triplet loss (Eq.~\eqref{eq:loss}), our method quickly learns to correctly separate matching from non-matching pairs and converges. As mentioned in Sec.~\ref{sec:method}, adding augmentations introduces harder samples and helps to produce more robust representations~\cite{augmentations_robustness}. We test applying the following set of augmentations: brightness and contrast alterations, color jittering, Gaussian blurring, and random cropping. The triplets selection is performed on the original frames to avoid altering the similarity values and thus reducing the quality of the sampled triplets. Therefore the augmented versions are used only when computing the loss.

We analyze different strategies to apply these augmentations (Tab.~\ref{tab:cyclecl_ablation_augmentations}): augmenting all samples, augmenting only the positives, or not using augmentations. We find that augmenting only the positives works best and it improves the \( F_1 \)-score by 7.5\% compared to not using augmentations. By augmenting only the positive samples and keeping the negatives unaltered, the model needs to learn to focus on the relevant information that encodes the current phase and ignore the changes introduced by the augmentations. Augmenting also the negatives makes the task again easier and performs worse than not using augmentations at all.
\begin{table}[t]
    \begin{center}
    \begin{tabular}{l|c|c}
    \hline
    \textbf{Data Aug.} & \textbf{AP} & \textbf{\( F_1 \)} \\
    \hline
      All & 50.19 & 51.98 \\
      Positives Only & 61.35 & 63.86 \\
      None & 55.64 & 56.39 \\
    \end{tabular}
    \end{center}
    \caption{Ablation study on the data augmentations for the Industrial dataset. Augmenting only the positives outperforms the other approaches by a significant margin.}
    \label{tab:cyclecl_ablation_augmentations}
\end{table}
\para{Sequence length.} The length of the sequence for sampling triplets is an important hyperparameter. During training, using a longer video sequence can provide more valid positive and negative pairs and improve the training signal by including more complex cases. Thus, performance is generally higher for longer input sequences (see Tab.~\ref{tab:cyclecl_ablation_seq_len}).
But there is a trade-off between sampling from long sequences and a reasonable batch size, due to memory constraints of the GPU. We see a small performance drop for a sequence of 200 frames with a batch size of 2. We believe that this is because there are too few samples for Batch Normalization to work well.
\subsection{Anomaly Detection}
\label{subsec:anomaly_detection}
We now evaluate how well the learned features transfer to an anomaly detection task.
\para{Evaluation protocol.}
We focus on the Industrial dataset and use the same ground truth as Sec.~\ref{sec:feature_quality}. Any interval annotated as non-periodic, which is rare, is considered as \emph{anomaly}. 
Anomalies are classified fully unsupervised based on the feature distance to other frames, in contrast to the previous section, where frames were classified based on the labels of other videos.
We use our frame embeddings as input to standard anomaly detection methods, namely \emph{\(k\)-Nearest Neighbors}, \emph{Local Outlier Factor} \cite{lof} and \emph{One-Class SVM} \cite{ocsvm}.
Using the available annotations we then compute the precision-recall curve and the corresponding AP. We also report the \emph{oracle \(F_1\)-score}, the maximum attainable \(F_1\)-score for the given anomaly ranking if the optimal classification threshold was chosen.

\para{Feature representation.} We use the learned features directly (\emph{Raw}). We also propose a version where a frame is represented by its similarity to other frames in the clip (\emph{Cycle}),~\ie by its row in the temporal similarity matrix, as \(c(v_i) = \mathcal{S}_{i:} \).

\para{Compared methods.}
As baseline, we apply the same techniques to features extracted from a model pre-trained on ImageNet, as done for the feature quality evaluation. We also compare our methods against an autoencoder (AE) and Mem-AE \cite{memae}, a state-of-the-art approach for anomaly detection based on reconstruction error. Since these methods directly provide an anomaly score, no additional post-processing techniques are needed.

\begin{table}[t]
    \begin{center}
    \begin{tabular}{l|l|c|c}
    \hline
    \textbf{Seq. Length} & \textbf{Batch Size} & \textbf{AP} & \textbf{\( k \)-NN \( F_1 \)} \\
    \hline
      {\color{gray} ImageNet} & {\color{gray} -} & {\color{gray} 46.25} & {\color{gray} 46.31} \\
      10 & 32 & 48.82 & 48.30 \\
      25 & 16 & 55.89 & 54.37 \\
      50 & 8 & 60.68 & 60.02 \\
      100 & 4 & 61.35 & 63.86 \\
      200 & 2 & 61.19 & 63.57 \\
    \end{tabular}
    \end{center}
    \caption{Ablation study on the sequence length of sampled video clips, for the Industrial dataset.}
    \label{tab:cyclecl_ablation_seq_len}
\end{table}

\para{Results.} Tab.~\ref{tab:anomaly_detection_results} shows the results. 
Our model outperforms all compared methods by a significant margin: it exceeds the ImageNet baseline by 13\% in \( F_1 \), even when using our cycle representation to improve it. The gap between the plain and memory-augmented autoencoders is even more extreme (13\% \( F_1 \) for Mem-AE \vs 53\% for our method). These methods do not perform well on this dataset, likely because the videos are visually rather static, and the high-capacity models used are able to learn to reconstruct the input successfully even in the presence of anomalies. The cycle representation provides significant improvements in the anomaly detection task compared to the raw embeddings  (with our method, from 28\% \( F_1 \) for raw to 53\% for cycle features).
This is because they capture a larger temporal context at a lower dimensionality, thus making it easier to represent changes in temporal dynamics.

\begin{table}[t]
    \begin{center}
    \begin{tabular}{l|l|l|c|c}
    \hline
      \textbf{Encoder} & \textbf{Features} & \textbf{AD Method} & \textbf{AP} & \textbf{ \( F_1 \)} \\
    \hline
      { ImageNet} & {Raw} & {LOF} & {12.8 } & { 19.8} \\
      { ImageNet} & {Cycle} & {\(k\)-NN } & { 26.2} & {39.8} \\
        AE & - & - & 08.6 & 09.3 \\
        MemAE \cite{memae} & - & - & 09.1 & 13.2 \\
        CycleCL & Raw & LOF & 19.4 & 27.9 \\
        CycleCL & Cycle & \(k\)-NN & 44.3 & 52.9 \\
    \end{tabular}
    \end{center}
    \caption{Comparison of different feature representations and anomaly detection methods according to the frame average precision (AP) and oracle \(F_1\) scores.
    }
    \label{tab:anomaly_detection_results}
\end{table}

\section{Conclusions}
\label{sec:conclusions}
In this work, we proposed CycleCL, a self-supervised learning method for periodic data. We designed a novel contrastive learning method based on a triplet loss. It iterates between (i) sampling frames from approximately the same phase as an anchor and negative frames from different phases and (ii) optimizing a feature encoder to increase their separation, until convergence.
We evaluated CycleCL on frame-based nearest-neighbour
classification and unsupervised anomaly detection on a new Industrial dataset and on three human actions datasets.
\new{
By achieving state-of-the-art results on such different domains, we demonstrate the effectiveness of our method on static and dynamic camera scenes, including both subtle and large visual differences.
}
Our extensive evaluation against the previous state of the art such as SimCLR~\cite{simclr}, RepNet~\cite{rep_dataset_countix}, Temporal Cycle Consistency~\cite{tcc} and DINO~\cite{dino} shows the benefits of our approach: it significantly outperforms all compared methods.

{\small
\bibliographystyle{ieee_fullname}
\bibliography{egbib}
}

\end{document}


\setcounter{page}{11}
\section{Supplementary Material}
\label{sec:supplementary}
We provide additional implementation details, ablations and additional visualizations in the following sections.
\setcounter{section}{0}
\renewcommand{\thesection}{\Alph{section}}  
\renewcommand{\thetable}{\Alph{table}}
\renewcommand{\thefigure}{\Alph{figure}}
\renewcommand{\theequation}{\Alpha{equation}}

\section{Implementation Details}
In addition to the pre-processing discussed in Sec. 4, each frame is resized to 224 x 224 and normalized before being fed to the network. The last convolutional layer produces a 7 x 7 x 512 feature maps for each frame, which are then stacked in chunks of 64 and forwarded to the projection head. We use 512 filters with \( 3 \times 3 \times 3 \) kernels for the 3D convolution, followed by batch normalization and Leaky ReLU activations. To maintain the input dimensionality on the temporal dimension, we set the padding to 1. The spatial dimensions are collapsed with either an adaptive max or average pooling.
After the final linear layer, the embeddings are \( L_2\) normalized. For optimization, we use Adam~\cite{adam_opt}, with a learning rate of \( 10^{-4} \) and weight decay of \( 10^{-3} \). The triplet loss margin \( \alpha \) is set to 0.5. When sampling the video clips, the temporal stride is set to 2 for the Industrial dataset and to 5 for the others. The experiments were run on an Nvidia Tesla T4 GPU.

\section{Additional Experiments}
This section discusses the additional experiments we performed to better understand the robustness of our method to varying hyperparameter choices.

\para{Embeddings dimensionality.} The dimensionality of the final output embedding can have a significant effect on the downstream task of nearest neighbor classification. In Tab.~\ref{tab:cyclecl_ablation_dimensionality} we compare different dimensionalities and show that they do not have a big influence on the results: overall our method performs best across ranges from 128 to 256. There seems to be a slight trade-off though: smaller embeddings suffer less from the curse of dimensionality when performing \(k\)-NN classification w.r.t. larger embeddings, but at the same time they have lower capacity to encode all the relevant information.
\begin{table}[h]
    \begin{center}
    \begin{tabular}{l|c|c}
    \hline
    \textbf{Output dim.} & \textbf{AP} &  \textbf{\( k \)-NN \( F_1 \)} \\
    \hline
      {\color{gray} ImageNet} & {\color{gray} 46.25} & {\color{gray} 46.31} \\
      64 & 58.13 & 58.55\\
      128 & 61.35 & 63.86 \\
      256 & 62.25 & 62.14 \\
    \end{tabular}
    \end{center}
    \caption{Ablation study on the output dimensionality, for the Industrial dataset.}
    \label{tab:cyclecl_ablation_dimensionality}
\end{table}

\para{Projection head.} We compare different head architectures besides the one presented in Sec.~3. In particular, we evaluate the contributions given by the 3D convolutional and the spatial max pooling layers w.r.t.~their~2D~and average counterparts.~Tab.~\ref{tab:cyclecl_ablation_head}~reports~the~results.~They indicate that both,~the~type~of~pooling~and~convolutional~layers, are important choices for maximizing performance.~The~3D~convolution~helps the model to capture temporal~context.~The spatial max pooling allows the model to focus on a specific region of the input and ignore background regions~that~have~no~information~regarding~the~cycle.

\begin{table}[h]
    \begin{center}
    \begin{tabular}{l|l|c|c}
    \hline
    \textbf{Layers} & \textbf{Pooling} & \textbf{AP} & \textbf{\( k \)-NN \( F_1 \)} \\
    \hline
      {\color{gray} ImageNet} & {\color{gray} -} & {\color{gray} 46.25} & {\color{gray} 46.31} \\
      2D Conv. + FC & Mean Pool & 51.71 & 54.32\\
      2D Conv. + FC & Max Pool & 56.09 & 57.73\\
      3D Conv. + FC & Mean Pool & 56.46 & 56.98 \\
      3D Conv. + FC & Max Pool & 61.35 & 63.86  \\
    \end{tabular}
    \end{center}
    \caption{Effect of the projection head on the performance, for the Industrial dataset.}
    \label{tab:cyclecl_ablation_head}
\end{table}

\para{\(L_2\)-normalization.} In the context of nearest neighbor classification, constraining the loss can help improve the quality of the embeddings \cite{knn_l2}. In particular, triplet loss is known to be sensitive to the magnitude of the embeddings \cite{triplet_loss}. Thus, we test the contribution of the \(L_2\) normalization bottleneck in stabilizing the training, reported in Tab.~\ref{tab:cyclecl_ablation_norm}.
In line with other self-supervised frameworks \cite{dino, simclr}, we find that \(L_2\)-normalization leads to significantly better performance.

\begin{table}[h]
    \begin{center}
    \begin{tabular}{l|c|c}
    \hline
    \textbf{Norm.} & \textbf{AP} & \textbf{\( k \)-NN \( F_1 \)} \\
    \hline
       {\color{gray} ImageNet} & {\color{gray} 46.25} & {\color{gray} 46.31} \\
      \( L_2 \)-norm & 61.35 & 63.86\\
      None & 57.13 & 56.01 \\
    \end{tabular}
    \end{center}
    \caption{Effect of  \( l_2 \)-normalization after the projection head, for the Industrial dataset.}
    \label{tab:cyclecl_ablation_norm}
\end{table}

\para{Data amount.} While our method requires no annotation, reducing the amount of data needed for training can greatly reduce training time. Therefore, we also perform an ablation study on the amount of data used to train the model, reported in Tab.~\ref{tab:cyclecl_ablation_data_amount}. While using more data can be effective in increasing the robustness of the learned representations, the improvements are minimal, and even using only 10\% of the data provides significant gains w.r.t. the ImageNet baseline.

\begin{table}[h]
    \begin{center}
    \begin{tabular}{l|l|c}
    \hline
    \textbf{Data \%} & \textbf{AP} & \textbf{\( k \)-NN \( F_1 \)} \\
    \hline
      {\color{gray} ImageNet} & {\color{gray} 46.25} & {\color{gray} 46.31} \\
      10 & 59.09 & 59.71 \\
      50 & 60.27 & 61.93 \\
      100 & 61.35 & 63.86 \\
    \end{tabular}
    \end{center}
    \caption{
Effect of the data quantity on the performance of our method, on the Industrial dataset.
}
\label{tab:cyclecl_ablation_data_amount}
\end{table}

\section{Additional Visualizations}

\begin{figure}[h]
    \centering
    \includegraphics[width=.98\linewidth]{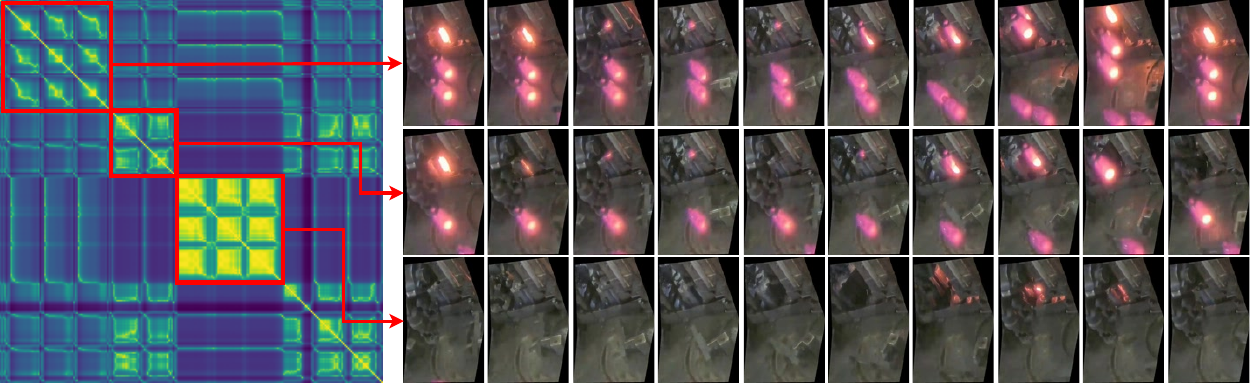}
    \caption{Analysis of the repeating patterns appearing in the TSM, extracted from an abnormal video. Three different phases can be identified (marked in red): (i) initial small anomaly, (ii) more severe anomaly, (iii) section running empty without producing any bottles.}
    \label{fig:tsm_examples_anomaly_with_frames}
\end{figure}

\para{Temporal self-similarity matrix (TSM).}
Here, we analyze the TSM shown in Fig.~4 of the main paper in more detail. The TSM shows the similarity of each frame to all (other) frames, thus encoding the temporal similarity patterns. We visualize the TSM and the corresponding frames for the anomalous video in Fig.~\ref{fig:tsm_examples_anomaly_with_frames}. From this, it can be seen that our CycleCL features clearly capture the difference in the cyclic signal when an anomaly occurs. In particular, it detects the different types of anomalies that occur in sequence as distinct clusters, represented by the four lighter blocks along the diagonal.

The first cluster depicts the start of an anomaly. At the beginning of the second cluster, the anomaly becomes more severe, and only one bottle is still produced. A cyclic pattern is still visible in both cases, but as the anomalies are different, the similarity between clusters is lower. During the third cluster, the bottle production is stopped entirely, while the robotic arm is still moving: the similarity is therefore very high within the cluster because most of the periodic process is stopped.

\begin{figure}[h]
    \centering
    \includegraphics[width=0.48\linewidth, height=0.12\linewidth]{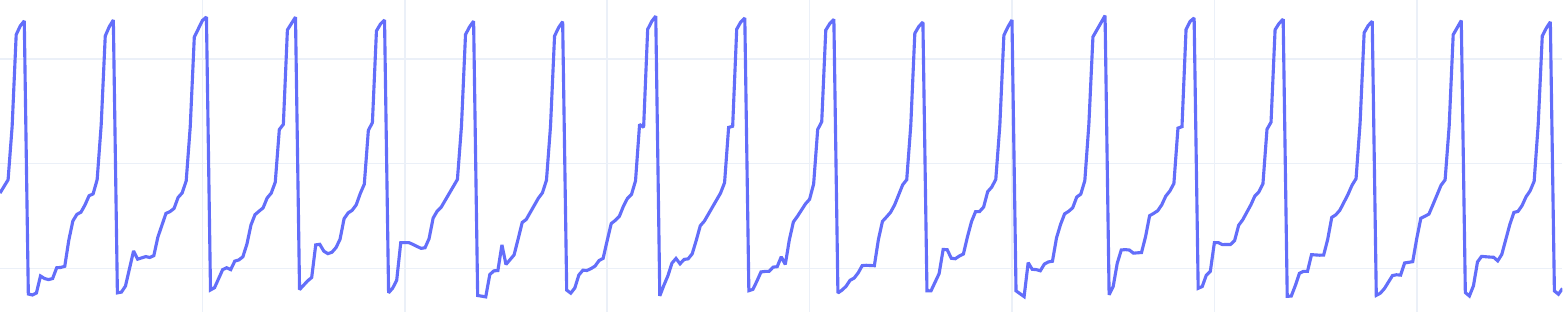}
    \hfill
    \includegraphics[width=0.48\linewidth, height=0.12\linewidth]{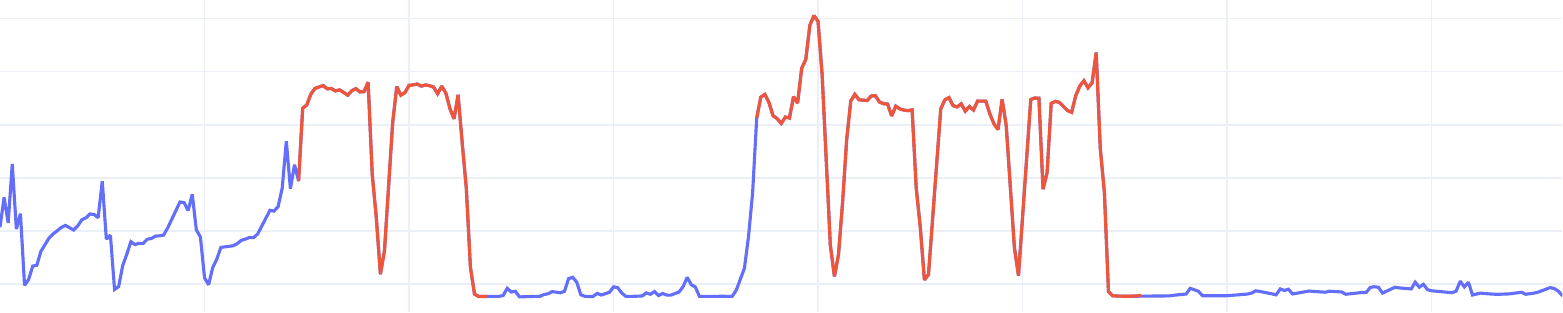}
    \caption{1D PCA projection of embeddings for a normal (left) and abnormal (right) video clip. The frames depicting the anomaly are represented in red.}
    \label{fig:pca_examples}
\end{figure}

\para{Features projection.} We apply PCA to the embeddings produced by the CycleCL model to map them to a 1-dimensional space, by taking the first principal component. Fig.~\ref{fig:pca_examples} shows the result of this operation for a normal and an anomalous video clip. The PCA projection of the normal video is smooth and quasi-sinusoidal, clearly showing the temporal cycle of the video, while the projection of the anomalous video is discontinuous.
This validates that our methods leads to features that are sensitive to the phase of the input.

\begin{figure}[h]
\vspace{4pt}
    \centering
    \includegraphics[width=1\linewidth]{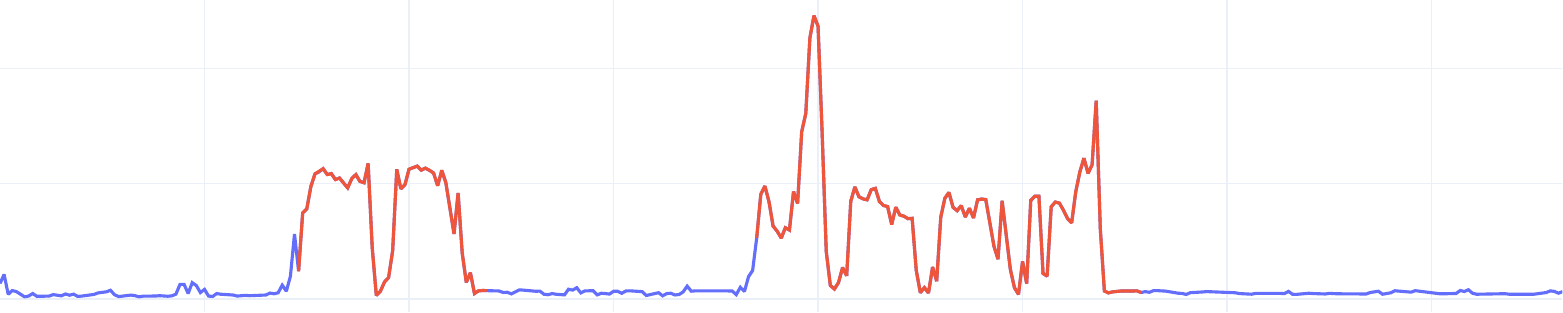}
    \caption{Nearest neighbor distances of abnormal frames (in red) and normal frames (in blue) with respect to a reference video depicting the normal process. While the process in the query video is running normally, correspondences with the reference can be found with a small distance. Instead, during an anomaly the visual difference is high and the distance to the nearest neighbor is a strong sign of abnormality.
    }
    \label{fig:1_nn_examples}
\end{figure}
\para{Nearest neighbor distance.} In the context of nearest neighbor classification, we are interested in understanding whether the model produces clearly separable embeddings for anomalous \emph{vs.} normal frames. Fig.~\ref{fig:1_nn_examples} shows an example of the nearest neighbor distances of an anomalous video \wrt{} a normal one. For each frame of the anomaly video, the distance to the nearest frame in the normal video is computed and used as its anomaly score. When a frame represents a normal state, a good match can be easily found in the normal video. On the contrary, when a frame represents an anomaly, the distance to the nearest neighbor is higher. This validates that the features are sensitive to deviations from normal repetitions.

{\small
\bibliographystyle{ieee_fullname}
\bibliography{egbib}
}